\documentclass[final]{cvpr}

\usepackage{times}
\usepackage{epsfig}
\usepackage{graphicx}
\usepackage{amsmath}
\usepackage{amssymb}

\usepackage{multirow}
\usepackage[pagebackref=true,breaklinks=true,colorlinks,bookmarks=false]{hyperref}

\begin{document}

\title{Dynamic Metric Learning: Towards a Scalable Metric Space \\
to Accommodate Multiple Semantic Scales}

\author{
Yifan Sun$^1$, Yuke Zhu$^1$, Yuhan Zhang$^2$, Pengkun Zheng$^1$, Xi Qiu$^1$, Chi Zhang$^1$, Yichen Wei$^1$\\
{$^1$Megvii Technology\hspace{0.5cm}}
{$^2$Beihang University}\\
{\texttt{\small{sunyf15@tsinghua.org.cn}} \hspace{0.5cm}}
{\texttt{\small{\{zhangchi,weiyichen\}@megvii.com}}}
}

\maketitle

\begin{abstract}
This paper introduces a new fundamental characteristic, \ie, the dynamic range, from real-world metric tools to deep visual recognition. In metrology, the dynamic range is a basic quality of a metric tool, indicating its flexibility to accommodate various scales. Larger dynamic range offers higher flexibility.  
In visual recognition, the multiple scale problem also exist. Different visual concepts may have different semantic scales. For example, ``Animal'' and ``Plants'' have a large semantic scale while ``Elk'' has a much smaller one.  Under a small semantic scale, two different elks may look quite \emph{different} to each other . However, under a large semantic scale (\eg, animals and plants),  these two elks should be measured as being \emph{similar}.  

Introducing the dynamic range to deep metric learning, we get a novel computer vision task, \ie, the Dynamic Metric Learning. It aims to learn a scalable metric space to accommodate  visual concepts across multiple semantic scales. Based on three types of images, \emph{i.e.},  vehicle, animal and online products, we construct three datasets for Dynamic Metric Learning. We benchmark these datasets with popular deep metric learning methods and find Dynamic Metric Learning to be very challenging. The major difficulty lies in a conflict between different scales: the discriminative ability under a small scale usually compromises the discriminative ability under a large one, and vice versa. As a minor contribution, we propose Cross-Scale Learning (CSL) to alleviate such conflict. We show that CSL consistently improves the baseline on all the three datasets. The datasets and the code will be publicly available at https://github.com/SupetZYK/DynamicMetricLearning.

\end{abstract}

\section{Introduction}

This papers consider the deep metric learning for visual recognition and supplements it with an important concept in metrology, \ie, the dynamic range. In metrology, the dynamic range is defined as the ratio between the largest and the smallest scale that a metric tool can provide. 
It is a basic quality of a metric, indicating the flexibility to accommodate various scales.  
We argue that such flexibility is also important for deep metric learning, because different visual concepts indeed correspond to different semantic scales. 
However, after a rethink on current deep metric learning tasks, we find that they all give NO consideration to the dynamic range.  
Therefore, we introduce the dynamic range to endow a single deep metric with flexibility among multiple semantic granularities. 
Potentially, it may reveal a new perspective in understanding the generalization ability of deep visual recognition.

\begin{figure}[t!]
	\centering
	\includegraphics[width=1\linewidth]{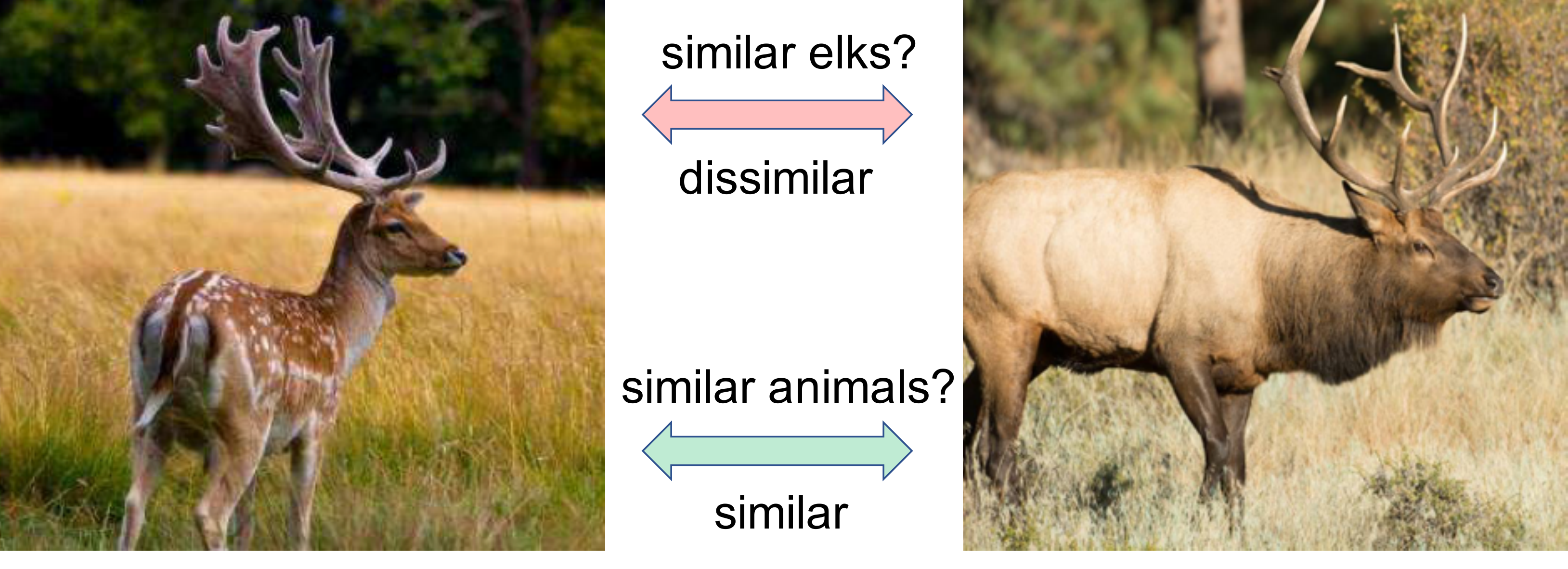}
	\caption{Visual concepts have various semantic scales, which impact on the similarity measuring result. Under the small scale of ``Elk'', these two elks look quite different. Under the large scale of ``Animal'', they should be measured as being similar.}
	\vspace{-4mm}
	\label{fig:elks}
\end{figure}

We explain the importance of ``dynamic range'' with a comparison between the deep metric learning and the real-world metric tools. In real world, a metric tool typically has a dynamic range. For example, a ruler has a lot of markings to indicate different lengths ranged from ``1 mm'' to ``10 cm'' or even to several meters. Rulers with only one single scale in ``1 mm'' or ``10 cm''  would have no use in daily life. Arguably, the dynamic range is essential to a metric tool, enabling it to measure objects of different sizes. In visual recognition, the visual concepts also have various semantic scales. For example,  ``Animals'' and ``Plants'' have a large semantic scale, while ``Elk'' has a much smaller one. When we try to measure the similarity between two images, which is the aim of metric learning, the underlying semantic scales impact on the result. In Fig. \ref{fig:elks}, the two elks look quite different to each other. However, under the large scale of ``Animals'', they should be judged as being similar. 

\begin{figure*}[t] 
\centering 
\includegraphics[width=0.95\textwidth]{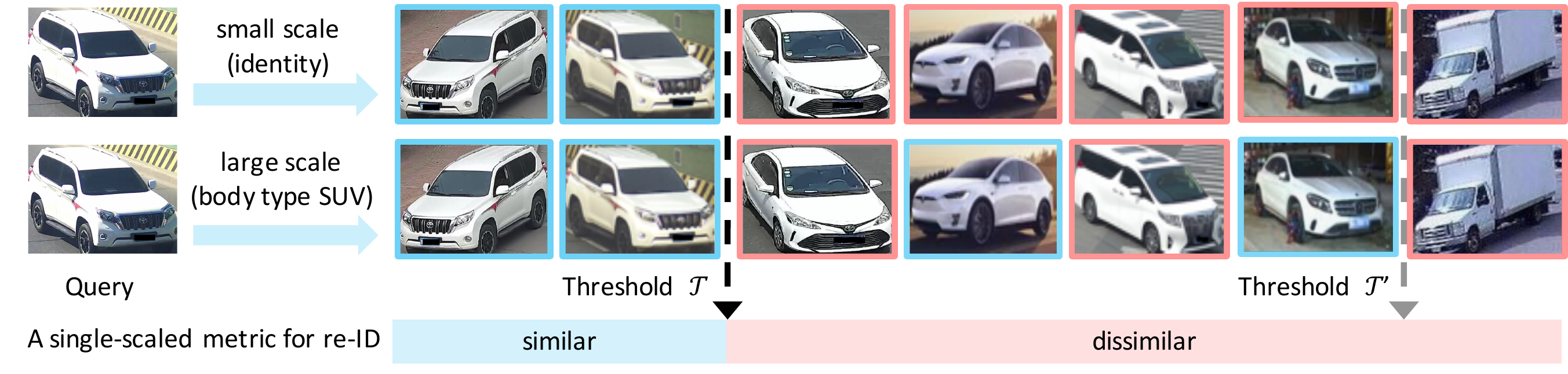} 
\caption{A single-scaled metric does not fit novel semantic scales. 
An accurate metric for vehicle re-ID (a small semantic scale) becomes inaccurate for recognizing the same body type ``SUV'' (\ie, a relatively larger semantic scale) in the second row. Using the threshold $\mathcal{T}$ fails to recall all the true matches, while using a lower threshold $\mathcal{T}$ incurs false positive matches. 
The \textcolor{red}{positive} and \textcolor{blue}{negative} matches are bounded with red and blue boxes, respectively. The images are from DyML-Vehicle.} 
\label{fig: intro} 
\end{figure*}

There is no consideration for the dynamic range in current deep metric learning tasks, \emph{e.g.}, face recognition \cite{wang2018cosface, wang2017normface, deng2019arcface, wang2018additive,wen2016discriminative,ranjan2017l2,liu2017sphereface,cao2018vggface2}, person re-identification \cite{Wang_2018MGN, Zheng_2019_CVPRJDGL, Sun_2018_ECCVPCB,Sun2019Perceive,8237672}, and vehicle re-identification (re-ID) \cite{lou2019embedding, he2019part, zhou2018vehicle, zhu2019vehicle}. They all focus on learning a metric for a single specified semantic scale (\eg, the identity of face, pedestrian and vehicles, respectively). The single-scaled metric lacks flexibility 
and may become inaccurate if the scale of interest changes. We validate this point with a toy scenario based on vehicle retrieval. In Fig. \ref{fig: intro}, two users use a same query image with different intentions. In the first row, the intention is to retrieve the cars with the same \emph{identity}, while the second row is to retrieve the cars with the same \emph{body type} (\ie, ``SUV''). A discriminative metric for vehicle re-ID (which learns to identify each vehicle) satisfies the first intention.  With a similarity threshold $\mathcal{T}$, it accurately separates the true matches and the false matches to the query image. However, it lacks discriminative ability for recognizing the same body type, which corresponds to a larger semantic scale than the identity.  Maintaining  $\mathcal{T}$ as the threshold, it fails to recall all the true matches. If we lower the similarity thresh hold to $\mathcal{T'}$ to promote the recall rate, the accuracy dramatically decreases (refer to  Section \ref{sec: single} for experimental evidence). We thus infer that a single-scaled metric does not fit novel semantic scales due to the lack of flexibility.

Introducing the dynamic range to deep metric learning, we get a new task, \emph{i.e.}, the \emph{Dynamic Metric Learning} (DyML). DyML aims to learn a scalable metric space to accommodate multiple semantic scales. In another word, a metric for DyML should be discriminative under several semantic granularities across a wide range.  To promote the research on DyML, we construct three datasets based on vehicle, animal and product, respectively. All these datasets have three different semantic scales, \ie, fine, middle and coarse. 
We benchmark these datasets with a variety of popular deep metric learning methods, \eg, Cosface \cite{cosface}, Circle Loss \cite{circleloss}, triplet loss \cite{schroff2015facenet}, N-pair loss \cite{Sohn2016ImprovedDM}. Extensive experiments show that DyML is very challenging. Even when the deep model learns from all the semantic scales in a multi-task manner, it does not naturally obtain a good dynamic range. The major difficulty lies in a conflict between different scales: the discriminative ability under a small scale usually compromises the discriminative ability under a large one, and vice versa. To alleviate such conflict, we device a simple method named \emph{Cross-Scale Learning} (CSL). CSL uses the within-class similarity of the smallest scale as the unique reference to contrast the between-class similarity of all the scales, simultaneously. Such learning manner is similar to the fact that all the markings on a ruler share ``0'' as the start. Experimental results confirm that CSL brings consistent improvement over the baselines. 


To sum up, this paper makes the following four contributions:

$\bullet$ We propose Dynamic Metric Learning by supplementing deep metric learning with dynamic range. In contrary to canonical metric learning for visual recognition, DyML desires discriminative ability across multiple semantic scales. 

$\bullet$ We construct three datasets for DyML, \ie, DyML-Vehicle, DyML-Animal and DyML-Product. All these datasets contain images under multiple semantic granularities for both training and testing. 

$\bullet$ We benchmark these DyML datasets with popular metric learning methods through extensive experiments. Experimental investigations show that DyML is very challenging due to a conflict between different semantic scales. 

$\bullet$ As a minor contribution, we propose Cross-Scale Learning for DyML. CSL gains better dynamic range and thus consistently improves the baseline.

\section{Related work}
\subsection{Deep Metric Learning.}
Deep metric learning (DML) plays a crucial role in a variety of computer vision applications, \eg, face recognition \cite{wang2018cosface, circleloss, deng2019arcface, wang2018additive,liu2017sphereface,cao2018vggface2}, person re-identification \cite{Wang_2018MGN, Zheng_2019_CVPRJDGL, Sun_2018_ECCVPCB,Sun2019Perceive,8237672}, vehicle re-identification  \cite{lou2019embedding, he2019part, zhou2018vehicle, zhu2019vehicle} and product recognition \cite{peng2020rp2k,bai2020products,goldman2019precise}. 
Generally, these tasks aim to retrieve all the most similar images to the query image. 

During recent years, there has been remarkable progresses \cite{wang2018cosface, wang2017normface, deng2019arcface, wang2018additive,wen2016discriminative,ranjan2017l2,liu2017sphereface,cao2018vggface2,circleloss} in deep metric learning. 
These methods are usually divided into two types, \ie, pair-based methods and classification-based methods. Pair-based methods (\eg, Triplet loss \cite{schroff2015facenet}, N-pair loss \cite{Sohn2016ImprovedDM}, Multi-Simi loss \cite{wang2019multi}) optimize the similarities between sample pairs in the deeply-embedded feature space. In contrast, classification-based methods learn the embedding by training a classification model on the training set, \eg, Cosface\cite{cosface}, ArcFace\cite{deng2019arcface}, NormSoftmax\cite{zhai2018classification} and proxy NCA\cite{movshovitz2017no}. 
Moreover, a very recent work, \ie, Circle Loss\cite{circleloss}, considers these two learning manners from a unified perspective. It provides a general loss function compatible to both pair-based and classification-based learning. 

Compared with previous metric learning researches, the Dynamic Metric Learning lays emphasis on the capacity to simultaneously accommodate multiple semantic scales. This new characteristic significantly challenges previous metric learning methods (as to be detailed in Section \ref{sec: exp_baseline}).

\subsection{Hierarchical Classification.}
We clarify the difference between DyML and a ``look-alike'' research area, \ie, the hierarchical classification \cite{tramer2020adaptive,khosla2011novel,berg2014birdsnap,krause20133d,liu2016deepfashion}. 
Dynamic Metric Learning organizes multiple semantic scales in a hierarchical manner (as to be detailed in Section \ref{sec: hierarchical}), which may seem similar to the hierarchical classification. However, DyML significantly differs from hierarchical classification in two major aspects. 

First, DyML belongs to metric learning domain, in which the training data and testing data has no class intersections. Correspondingly, the learned metric has to be generalized to unseen classes. In contrast, the hierarchical classification belongs to image classification task. The training data and the testing data share same classes, so there is no unseen classes during testing. 

Second, DyML uses the hierarchical information for learning a single hidden layer (\ie, the deep embedding layer). All the semantic scales (coarse, medium and fine levels) are equally important for DyML. In contrast, the hierarchical classification methods mainly cares about the accuracy on the fine level, and all the other semantic scales are used only for auxiliary supervision (in preceding layers before the final classification layer). 

Moreover, the hierarchical data structure is NOT prerequisite for DyML. We organize the multiple semantic scales in hierarchy mainly for efficiency consideration.


\section{Dynamic Metric Learning} \label{sec: problem}

\subsection{Task Formulation}
Let us assume there are $C$ categories of images $\{I_1, I_2, \cdots, I_C\}$. Each $I_i$ is consisted of several images. Given a random category $I_i$, if there exists another category $I_{j \ne i}$ looks very similar to $I_i$, we consider these categories (as well as the corresponding images) jointly form a small semantic scale. In contrast, given a random category $I_i$, if its nearest neighbor $I_{j \ne i}$ looks quite different from $I_i$, we consider these categories jointly form a large semantic scale. To be intuitive, we take the animals as an example. According to biological taxonomy, animals may be divided by ``phylum'', ``class'', ``order'', ``family'', ``genus'', \etc. The categories in the ``phylum'' form a large semantic scale, while the categories in ``genus'' form a relatively small one. 

Given multiple semantic scales $\mathcal{S}^1, \mathcal{S}^2, \cdots, \mathcal{S}^M$, DyML ensembles all of them to expand a wide semantic range $\mathbb{R}$, which is formulated by:
\begin{equation}
\mathbb{R} = \{\mathcal{S}^1, \mathcal{S}^2, \cdots, \mathcal{S}^M\},
\end{equation}
in which a random semantic scale $\mathcal{S}^i$ contains $N^i$ labeled images, \ie, $\{(x_j^i, l_j^i)\}$ ($j=1, 2, \cdots, N^i$).  

DyML aims to learn a single metric space $\mathcal{M}$ with discriminative ability across the whole semantic range $\mathbb{R}$. Concretely, through $\mathcal{M}$, samples of a same class are close to each other, and the samples of different classes are far away, regardless of the underlying semantic scale $\mathcal{S}^i$.  
To evaluate the discriminative ability under each scale, DyML adopts the image retrieval paradigm. Given a query image $x_q^i$ from $\mathcal{S}^i$, DyML employs $\mathcal{M}$ to calculate its similarity scores between all the other images $x_{j\neq q}^i$. According to descending order of the similairty scores, DyML get a ranking list, denoted as $\{r_1^i, r_2^i, \cdots, r_{N_i}^i\}$, where $r_{j\neq q}^i$ is the sorted index of image $x_{j\neq q}^i$. An ideal ranking list is to place all the positive images (\emph{i.e.}, images from a same class) in front of the negative images. In another word, $r_j^i$ should be small if $x_j^i$ and $x_q^i$ are within a same class (\emph{i.e.}, $l_j^i=l_q^i$). Formally, the objective function of DyML is formulated as,
\begin{equation}
    \min\sum_{j=1}^{N_{i}}r_{j\neq q}^i \mathcal{I}(l_j^i, l_p^i), \qquad \forall \mathcal{S}^i \in \mathbb{R}
\end{equation}
in which $\mathcal{I}(l_j^i, l_p^i)=1$ if $l_j^i=l_q^i$ and $\mathcal{I}(l_j^i, l_p^i)=0$ if $l_j^i\neq l_q^i$.


\begin{figure*}[t] 
\centering 
\includegraphics[width=1.0\textwidth]{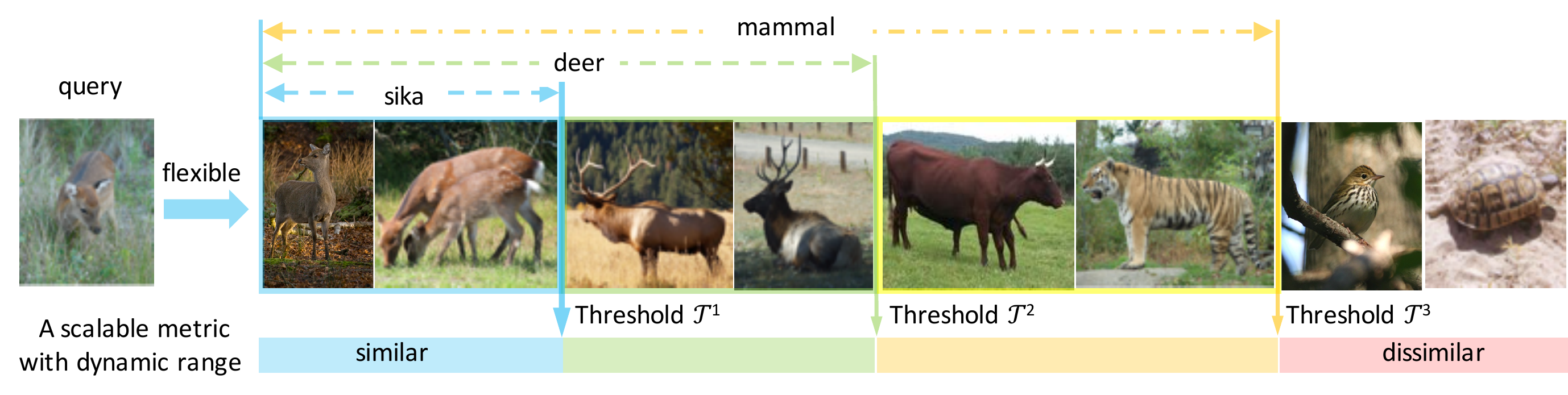} 
\vspace{-1em}\caption{A multi-scaled metric with a dynamic range. It is capable to accommodate different semantic scales. Given a ``sika'' as the query image, it retrieves images of ``sika'', ``deer'' and ``mammal'' with descending similarity thresholds, \ie, $\mathcal{T}^1 > \mathcal{T}^2 > \cdots > \mathcal{T}^M$. The images are from DyML-Animal.} 
\label{fig: metric} 
\end{figure*}

\subsection{Hierarchical Modification for Efficiency} \label{sec: hierarchical}
We note that multiple semantic scales may share images with each other to reduce the cost for image collection. In another word, a single image $x$ may simultaneously belong to  multiple $\mathcal{S}$ (with the superscript omitted). It is because an image in itself may have several visual concepts. For example, an image of elk (in Fig. \ref{fig:elks}) may  correspond to ``Elk'' (small scale), ``Deer'' (medium scale) or ``Mammal'' (large scale). 
In practice, when preparing the dataset (as to be detailed in Section \ref{sec: dataset}), we annotate each image $x_j$ with a set of labels $L_j$, which is formulated as:
\begin{equation}
    \mathbb{R}= \{(x_1, \mathcal{L}_1),(x_2, \mathcal{L}_2), \cdots, (x_N, \mathcal{L}_N)\},
\end{equation}
in which N is the total number of images, $\mathcal{L}_j = \{l_j^1, l_j^2, \cdots, l_j^M\}$ is the complete label set for $x_j$. Specifically, $l_j^i$ denotes the label of $x_j$ under semantic scale $\mathcal{S}_i$. 
$\mathcal{S}^1, \cdots, \mathcal{S}^M$ have gradually ascending scales, \ie, $\mathcal{S}^1 \subset \mathcal{S}^2 \subset \cdots \subset \mathcal{S}^M$. Multiple classes under $\mathcal{S}^{i}$ may belong to a same large-scale class in $\mathcal{S}^{j}(j>i)$. Consequentially, the semantic scales of DyML follow a hierarchical ordering. 


\subsection{A Multi-scaled Metric for DyML}
Under the hierarchical modification, \ie, $\mathcal{S}^1 \subset \mathcal{S}^2 \subset \cdots \subset \mathcal{S}^M$, we analyze the property of a multi-scaled metric $\mathcal{M}$. To be accurate under all the scales, $\mathcal{M}$ should satisfy the following two criteria:

$\bullet$ \textbf{Single-scale criterion.} Under a same scale, $\mathcal{M}$ maintains within-class compactness, as well as between-class discrepancy, \ie, the within-class similarity scores are larger than the between-class similarity scores. It is a common criterion in all metric learning tasks. 

$\bullet$ \textbf{Cross-scale criterion.} Second, under any two different scales $\mathcal{S}^i \subset \mathcal{S}^j$, the within-class similarity scores under $\mathcal{S}^i$ should be larger than that under $\mathcal{S}^j$. It is a unique criterion in Dynamic Metric Learning. 

With both the single-scale criterion and the cross-scale criterion satisfied, $\mathcal{M}$ uses descending thresholds (\ie, $\mathcal{T}^1 > \mathcal{T}^2 > \cdots > \mathcal{T}^M$) to recall true matches under ascending semantic scales (\ie, $\mathcal{S}^1 \subset \mathcal{S}^2 \subset \cdots \subset \mathcal{S}^M$), as illustrated in Fig. \ref{fig: metric}. Given an image of ``sika deer'' as the query, the gallery images of ``sika deer'', ``deer (but not sika)'', ``mammal (but not deer)'', ``bird'', ``reptile'' are recognized as having gradually-decreasing similarities. It thus looks like a ruler with multiple markings to accommodate objects with various sizes. 


\section{DyML Datasets}\label{sec: dataset}

\subsection{Description} 

\textbf{Overview.} This paper provides three datasets for Dynamic Metric Learning research, \ie, DyML-Vehicle, DyML-Animal and DyML-Product. We collect all the source images from publicly available datasets and supplement them with some manual annotations to enrich the semantic scales. Overall, these datasets have the following common properties:

    $\bullet$\emph{Three hierarchical levels of labelling}. Each image may have at most three labels corresponding to coarse, middle and fine levels, respectively. A coarse class contains several middle classes. Similarly, a middle class contains several fine classes. 
    
    $\bullet$ \emph{Abundant semantic scales}. Although the datasets are hierarchically organized into three levels, the actual semantic scales are even more abundant. It is because each level may contain several semantic scales. For example, in DyML-Animal, the visual concepts contained in the middle level are consisted of ``order'', ``family'' and ``genus''. 
    
    $\bullet$ \emph{No class intersections between training and testing set}. In accordance to the popular metric learning settings, the testing classes are novel to training classes (except for the coarse level). The total classes under coarse level are very limited. To obtain enough training / testing classes, we allow intersections under the coarse level, and insist open-set setting. In another word, under the coarse level, some testing classes exist in the training set, while some other testing classes are novel.

\begin{table*}[t]
\setlength{\tabcolsep}{15pt}
\footnotesize
\centering
\begin{tabular}{c|c|rr|rr|rr}
\hline
\multicolumn{2}{c|}{\multirow{2}{*}{Datasets}} & \multicolumn{2}{c|}{DyML-Vehicle} & \multicolumn{2}{c|}{DyML-Animal} & \multicolumn{2}{c}{DyML-Product} \\ \cline{3-8} 
\multicolumn{2}{c|}{}      & Train            & Test           & Train           & Test           & Train            & Test           \\ \hline \hline
\multirow{2}{*}{Coarse}    & Classes    & 5                & 6              & 5               & 5              & 36               & 6              \\ 
                           & Images     & 343.1 K          & 5.9 K          & 407.8 K         & 12.5 K         & 747.1 K          & 1.5 K          \\ \hline
\multirow{2}{*}{Middle}    & Classes    & 89               & 127            & 28              & 17             & 169              & 37             \\ 
                           & Images     & 343.1 K          & 34.3 K         & 407.8 K         & 23.1 K         & 747.1 K          & 1.5 K          \\ \hline
\multirow{2}{*}{Fine}      & Classes    & 36,301           & 8,183          & 495             & 162            & 1,609            & 315            \\ 
                           & Images     & 343.1 K          & 63.5 K         & 407.8 K         & 11.3 K         & 747.1 K          & 1.5 K          \\ \hline
\end{tabular}
\caption{Three datasets, \ie, DyML-Vehicle, DyML-Animal, DyML-Product for Dynamic Metric Learning. We collect the raw images from publicly available datasets and supplement them with abundant multi-scale annotations. Each dataset has three hierarchical labels ranging from coarse to fine. Some level contains several semantic scales. Under the middle level and fine level, there is no intersection between training and testing classes. The coarse level allows certain class intersections and yet insists on the open-set setting.}
\label{tab: dataset}
\end{table*}

The quantitative descriptions of all these three datasets are summarized in Table \ref{tab: dataset}.

\textbf{DyML-Vehicle} merges two vehicle re-ID datasets PKU VehicleID~\cite{pku_vehicle}, VERI-Wild~\cite{veri-wild}. Since these two datasets have only annotations on the identity (fine) level, we manually annotate each image with ``model" label (\eg, Toyota Camry, Honda Accord, Audi A4) and ``body type" label (\eg, car, suv, microbus, pickup). Moreover, we label all the taxi images as a novel testing class under coarse level.   

\textbf{DyML-Animal} is based on animal images selected from ImageNet-5K~\cite{imagenet1}. It has 5 semantic scales (\ie, classes, order, family, genus, species) according to biological taxonomy. Specifically, there are 611 ``species'' for the fine level, 47 categories corresponding to ``order'', ``family'' or ``genus'' for the middle level, and 5 ``classes'' for the coarse level. 
We note some animals have contradiction between visual perception and biological taxonomy, \eg, whale in ``mammal'' actually looks more similar to fish. Annotating the whale images as belonging to mammal would cause confusion to visual recognition. So we take a detailed check on potential contradictions and intentionally leave out those animals. 

\textbf{DyML-Product} is derived from iMaterialist-2019 \footnote{https://github.com/MalongTech/imaterialist-product-2019}, a hierarchical online product dataset. The original iMaterialist-2019 offers up to 4 levels of hierarchical annotations. We remove the coarsest level and maintain 3 levels for DyML-Product.

\subsection{Evaluation Protocol}
\textbf{The overall CMC and mAP.}
DyML sets up the evaluation protocol based on two popular protocols adopted by image retrieval, \ie, the Cumulated Matching Characteristics (CMC) \cite{Wang2007Shape} and the mean Average Precision (mAP) \cite{market} The criterion of CMC indicates the probability that a true match exists in the top-K sorted list. In contrast, the criterion of mAP considers both precision and recall of the retrieval result. When there are multiple ground-truth for a query (which is the common case), mAP lays emphasis on the capacity of recognizing all the positive matches, especially those difficult ones. 

To get an \textbf{overall evaluation} on the discriminative ability under all the semantic scales, DyML first evaluates the performance under each level and then averages the results under three levels (\ie, fine, middle and coarse). Notablly, the level information is not accessible to the evaluated metric. Manually using the level information of the query to fit the underlying scale is not allowed. It thus prohibits learning several single-scaled metrics and manually choosing an appropriate one to fit each query. The reason is that, in reality, 1) user will not know which metric exactly fits the underlying scale and 2) enumerating all the metrics online is impractical. 

\textbf{The average set intersection (ASI).} ASI is a popular protocol for evaluating the similarity of two ranking list. Given two ranking list $A=\{a_1, a_2, \cdots, a_N\}$ and $B=\{b_1, b_2, \cdots, b_N\}$, the set intersection at depth $k$ is defined as:
\begin{equation}
    SI(k)= \frac{|\{a_1, a_2, \cdots, a_k\} \cap \{b_1, b_2, \cdots, b_k\}|}{k},
\end{equation}
in which $|\bullet|$ denotes the operation of counting the number of a set.

ASI averages SI at random depths by:
\begin{equation}
    ASI= \frac{1}{N}\sum_{i=1}^N SI(i)
\end{equation}

In DyML, we use the ground truth ranking list and the predicted ranking list to calculate ASI. ASI naturally takes all the semantic scales into account.

\section{Methods}
\subsection{Multi-scale learning Baseline}
\label{subsec:baselines}


Basically, we use a deep model (backboned on ResNet-34 \cite{resnet}) to map the raw input images into a feature space. 
Given the deep features, we first enforce a independent supervision through a specified loss function (\eg, the softmax loss, Cosface \cite{cosface}, Circle Loss \cite{circleloss}, Triplet loss \cite{schroff2015facenet}, N-pair loss \cite{Sohn2016ImprovedDM}, Multi-Simi loss \cite{wang2019multi}). Then we sum up the losses on all the semantic scales in the multi-task learning manner. The multi-scale learning baseline has the following characteristics:

$\bullet$ First, it is superior to the single-scaled metric learning \wrt to the overall accuracy. Since it combines the supervisions under multiple semantic scales, the improvement on the overall accuracy is natural. The details are to be accessed in Section \ref{sec: single}.

$\bullet$ Second, it is confronted with a mutual conflict among different scales. To illustrate this point, let us assume two samples $x_1$ and $x_2$ with $l_1^i \ne l_2^i$ and $l_1^{i+1} = l_2^{i+1}$. In another word, under the small scale $\mathcal{S}^i$, they belong to two different classes, while under the larger scale $\mathcal{S}^{i+1}$, they belong to a same class. Under $\mathcal{S}^i$, the baseline is to push $x_1$ and $x_2$ far away. In contrast, under $\mathcal{S}^{i+1}$, the baseline is to pull them close. These two contrary optimization objectives compromise each other. In Section \ref{sec: exp_mutual}, we experimentally validate the above-described mutual conflict.

\subsection{Our Method: Cross-Scale Learning}

In response to the mutual conflict, we propose Cross-Scale Learning (CSL). 
Under $\mathcal{S}^i$, we denote the within-class similarity as $s_p^i$, and the between-class similarity as $s_n^i$. 
CSL uses the within-class similarity under the smallest scale (\ie, $s_p^1$) as the unique reference to contrast the between-class similarities under all the scales (\ie, $s_n^i, i=1, 2, \cdots, M$), yielding the so-called Cross-Scale Learning. Formally, CSL desires: 

\begin{equation}\label{eq: csl1}
    s_p^1 \ge s_n^i + m^i,  \qquad i=1,2,\cdots, M
\end{equation}
in which $m^i$ is the similarity margin under $\mathcal{S}^i$. Intuitively, we set $m^1<m^2<\cdots<m^M$ in accordance to the increasing scope of $\mathcal{S}^1, \mathcal{S}^2,\cdots, \mathcal{S}^M$. 

Eq. \ref{eq: csl1} enables a joint optimization across all the semantic scales, with $\mathcal{S}^1$ as the reference scale. 
The advantages of CSL are two-fold. First, it does not enforce explicit within-class compactness under $\mathcal{S}^i (i>1)$, and thus avoids the mutual conflict between different scales. Second, using $\mathcal{S}^1$ as the shared reference scale (for optimizing the between-class similarities under all the scales) makes the learning more stable. In reality, a ruler has all its markings annotated with distances from the ``0'' point. 

The loss function for Cross-Scale Learning is correspondingly defined as:
\begin{equation}
\mathcal{L}_{CSL}=\sum_{i=1}^M{\log(1+\sum_{k=1}^{C^i}\exp{\alpha(s_{n,k}^i - s_p^1+m^i)})},
\end{equation}\label{eq: csl2}
in which $\alpha$ is a scaling factor, $C^i$ is the total number of training classes under $\mathcal{S^i}$, $s_{n,k}^i$ is the $k$-th between-class similarity under $\mathcal{S}^i$.

Besides the cross-scale optimization, CSL further eliminates the mutual suppression with a \emph{proxy-sharing strategy}. Basically, CSL adopts the classification-based training manner: The within-class similarity $s_p$ is calculated as the cosine similarity between feature $x$ and the weight vector of its target class. Meanwhile, a between-class similarity $s_n$ is calculated as the cosine similarity between feature $x$ and a corresponding weight vector of a non-target class. In CLS, only the classes under $\mathcal{S}^1$ has an independent weight vector as the class prototype. A high-level class under the higher semantic scale $\mathcal{S}^i(i>1)$ uses the set of weight vectors consisted of its sub-classes in $\mathcal{S}^1$ as the prototype. It thus allows a 
high-level class to have relatively large within-class diversity \cite{Qian_2019_ICCV}. To measure the between-class similarity between $x$ and a high-level class, CSL has to compare $x$ against a set of weight vectors. To this end, we follow the common practice of hard mining strategy, \ie, choosing the hardest (closest) negative weight vector against $x$ to represent the whole vector set. 

\section{Experiments}
The experiments are arranged as follows. Section \ref{sec: single} first experimentally validates that a single-scaled metric does not fit novel semantic scales. Section \ref{sec: exp_baseline} benchmarks the DyML datasets with popular metric learning methods. Section \ref{sec: exp_CSL} evaluates the proposed method of Cross-Scale Learning. Section \ref{sec: exp_mutual} experimentally validates that CSL alleviates the mutual conflict among different scales. 

\subsection{Analysis on Single-scaled Metric} \label{sec: single}

We investigate the generalization capacity of single-scaled metrics on DyML-Vehicle. Specifically, we learn a deep metric using only the coarse-level, middle-level and the fine-level labels, respectively. We compare them against the multi-scale learning baseline in Fig. \ref{fig: single_multi}. For fair comparison, we use a same loss function, \ie, Cosface \cite{cosface} under all the settings. We draw three observations as follows:

\begin{figure}[t!]
	\centering
	\includegraphics[width=1\linewidth]{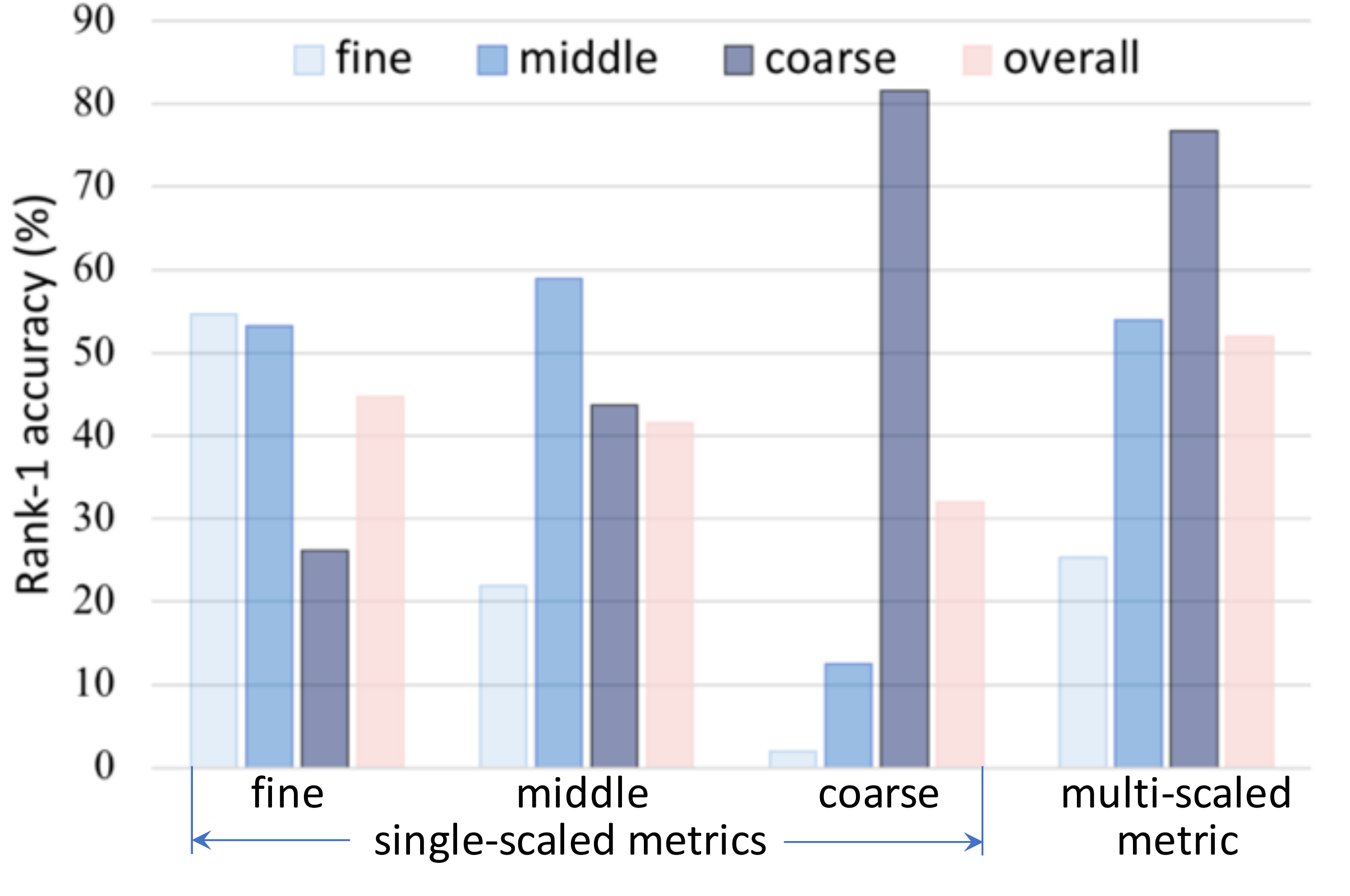}
	\caption{Comparison between single-scaled metrics and a multi-scaled metric baseline on DyML-Animal. The multi-scaled metric surpasses all the single-scaled metrics on the overall accuracy.}
	\vspace{-4mm}
	\label{fig: single_multi}
\end{figure}

\begin{table*}[]
\centering
\label{tab:my-table}
\resizebox{\textwidth}{!}{%
\begin{tabular}{llllll|lllll|lllll}
\hline
\multicolumn{1}{c}{} & \multicolumn{5}{c|}{DyML-Vehicle} & \multicolumn{5}{c|}{DyML-Animal} & \multicolumn{5}{c}{DyML-Product} \\ \cline{2-16} 
                     & ASI  & mAP  & R@1 & R@10 & R@20 & ASI  & mAP  & R@1 & R@10 & R@20 & ASI  & mAP  & R@1  & R@10 & R@20 \\ \hline
Triplet Loss         & 18.3      & 10.0    & 13.8   & 52.6   & 65.1    & 19.3 & 11.0  & 18.2 & 55.5 & 66.3 & 9.2   & 9.3  & 11.2 & 43.6 & 53.3 \\
MS Loss              & 19.7      & 10.4    & 17.4   & 56.0    & 67.9    & 19.9 & 11.6 & 16.7 & 53.5 & 64.8 & 9.8   & 10.0 & 12.7 & 45.7 & 56.4 \\
N-Pair Loss          & 19.4       & 10.5    & 16.4   & 55.7    & 68.1    & \textbf{45.7} & 30.3 & 39.6 & 69.6 & 78.8 & 15.7  & 15.3 & 20.3 & 55.5 & 65.6  \\
\hline
Softmax Loss         & 22.7     & 12.0    & 22.9   & 61.6    & 72.9    & 35.9 & 25.8 & 49.6 & 81.7 & 88.8  & 26.8  & 26.1 & 50.2 & 81.6 & 87.7 \\
Cosface Loss         & 23.0      & 12.0    & 22.9   & 62.1    & 73.4    & 39.6 & 28.4 & 45.1 & 75.7 & 83.3 & 25.5  & 25.0 & 49.3 & 81.3 & 87.7 \\
Circle Loss          & 22.8      & 12.1    & 23.5   & 62.0    & 73.3    & 44.5 & 30.6 & 41.5 & 72.2 & 80.3 & 15.8  & 15.0 & 26.7 & 61.5 & 70.3 \\
\hline
CSL                  & \textbf{23.0}    & \textbf{12.1}    & \textbf{25.2}   & \textbf{64.2}    & \textbf{75.0}     & 45.2  & \textbf{31.0} & \textbf{52.3} & \textbf{81.7} & \textbf{88.3}    & \textbf{29.0}  & \textbf{28.7} & \textbf{54.3} &\textbf{83.1} & \textbf{89.4} \\ \hline
\end{tabular}%
}
\caption{Evaluation of six popular deep learning methods and the proposed Cross-Scale Learning (CSL) on DyML-Vehicle, DyML-Animal and DyML-Product. For CMC and mAP, we report the overall results averaged from three scales. The ASI is an overall evaluation protocol in its nature. Best results are in \textbf{bold}.}
\label{tab: evaluation}
\end{table*}

First, each single-scaled metric shows relatively high accuracy under its dedicated scales. For example, the ``fine'' metric (\ie, the metric learned with fine-level labels) achieves 54.7\% Rank-1 accuracy under the fine-level testing. Second, a single-scaled metric does not naturally generalize well to another scale. For example, under the coarse-level testing, the ``fine'' metric only achieves 26.2\% Rank-1, which is lower than the ``coarse'' metric by $-55.4\%$. It validates that a single-scaled metric does not fit a novel scale. Third, with consideration of the overall performance, we find that multi-scale training performs the best. Under the overall evaluation, it surpasses the ``fine'' metric, the ``middle'' metric and the ``coarse'' metric by +7.2\%, + 10.4\%, + 19.9\%, respectively. It implies that multi-scale training benefits from the multi-scale information and thus improves the overall performance. 

Since the objective of DyML is the discriminative ability under all the semantic scales, we use multi-scale supervision for all the evaluated  baselines to promote the overall performance.

\subsection{Methods Evaluation} \label{sec: exp_baseline}

We benchmark all the DyML-datasets with 6 popular deep metric learning methods, including 3 pair-based methods (\ie, the Triplet loss \cite{schroff2015facenet}, N-pair loss \cite{Sohn2016ImprovedDM}, Multi-Simi loss \cite{wang2019multi}) and 3 classification-based methods (\ie, the Softmax Loss, Cosface \cite{cosface} and Circle Loss \cite{circleloss}). For each method, we use multi-scale supervision for training. For evaluation with CMC and mAP, we average the performance under all the scales and only report the overall performance. The results are reported in Table \ref{tab: evaluation}, from which we draw three observations. 

First, DyML is very challenging. The overall performance is low under all the three baselines. For example, the ``Cosface loss'' only achieves 23.0\%, 39.6\%, 25.5\% ASI and 12.0\%, 30.7\% and 25.0\% mAP on DyML-Vehicle, DyML-Animal and DyML-Product, respectively. Second, the classification-based methods generally surpasses the pair-based methods, indicating that the classification training manner usually achieves higher discriminative ability. It is consistent with the observation in many other metric learning tasks \cite{wang2018cosface, wang2018additive,wen2016discriminative}. We infer that in spite of the fundamental difference of dynamic range, DyML shares a lot of common properties with the canonical deep metric learning. Third, comparing three  classification-based methods (\ie, softmax, Cosface and Circle Loss) against each other, we find that they all achieve very close performance. Though Cosface and Circle Loss marginally surpasses the softmax loss in 
canonical deep metric learning tasks \cite{wang2018cosface, wang2018additive, circleloss}, they do NOT exhibit obvious superiority for DyML. One potential reason is that Cosface and Circle Loss has more hyper-parameters (\ie, the scale and the margin) for each semantic scale. DyML has multiple semantic scales and thus make the optimization of these hyper-parameters more difficult.

\begin{table}[t]
\small
\center
\setlength{\tabcolsep}{7pt}
\begin{tabular}{llcccc}
\hline
\multirow{2}{*}{Method}  & \multirow{2}{*}{Scale} & \multicolumn{2}{c}{DyML-Animal} & \multicolumn{2}{c}{DyML-Product} \\ \cline{3-6} 
                         &                        & mAP            & R@1            & mAP             & R@1            \\ \hline
\multirow{4}{*}{Cosface} & Fine                   & 8.7            & 18.3           & 11.1            & 20.3           \\
                         & Middle                 & 28.4           & 46.6           & 16.9            & 47.6           \\
                         & Coarse                 & 48.2           & 70.5           & 47.1            & 80.0           \\
                         & Overall                & 28.4           & 45.1           & 25.0            & 49.3           \\ \hline
\multirow{4}{*}{CSL}     & Fine                   & 10.3           & 25.3           & 15.6            & 26.2           \\
                         & Middle                 & 30.1           & 53.9           & 20.1            & 53.2           \\
                         & Coarse                 & 52.7           & 77.7           & 50.4            & 83.7           \\
                         & Overall                & 31.0           & 52.3           & 28.7            & 54.3           \\ \hline
\end{tabular}
\caption{Comparison between Cosface and the proposed CSL in three specified scales (besides the overall performance). We report mAP and Rank-1 accuracy. CSL exhibits consistent improvement under all the scales.}
\label{tab: level_cmp}
\end{table}


\subsection{Effectiveness of Cross-Scale Learning} \label{sec: exp_CSL}

We compare the proposed method, \ie, Cross-Scale Learning with all the six existing methods in Table \ref{tab: evaluation}. It clearly shows that CSL is superior to the competing methods \wrt the overall performance. For example, CSL surpasses ``Cosface'' by \textcolor{black}{+2.3\%}, \textcolor{black}{+7.2\%} and \textcolor{black}{+5.0\%} Rank-1 accuracy on DyML-Vehicle, DyML-Animal and DyML-Product, respectively. 
To be more concrete, Table \ref{tab: level_cmp} compares CSL against Cosface under each (fine, middle and coarse) semantic scale, respectively. We observe that  CSL achieves improvement not only on the overall (averaged) accuracy, but also on every single scale level. It indicates that CSL does not have a bias towards certain specified semantic scales. Instead, it generally improves the discriminative ability of the learned deep metric under (almost) all the scales, indicating better generalization across multiple scales.

\subsection{Reasons for the Superiority of CSL} \label{sec: exp_mutual}

\begin{figure}[t!]
	\centering
	\includegraphics[width=1\linewidth]{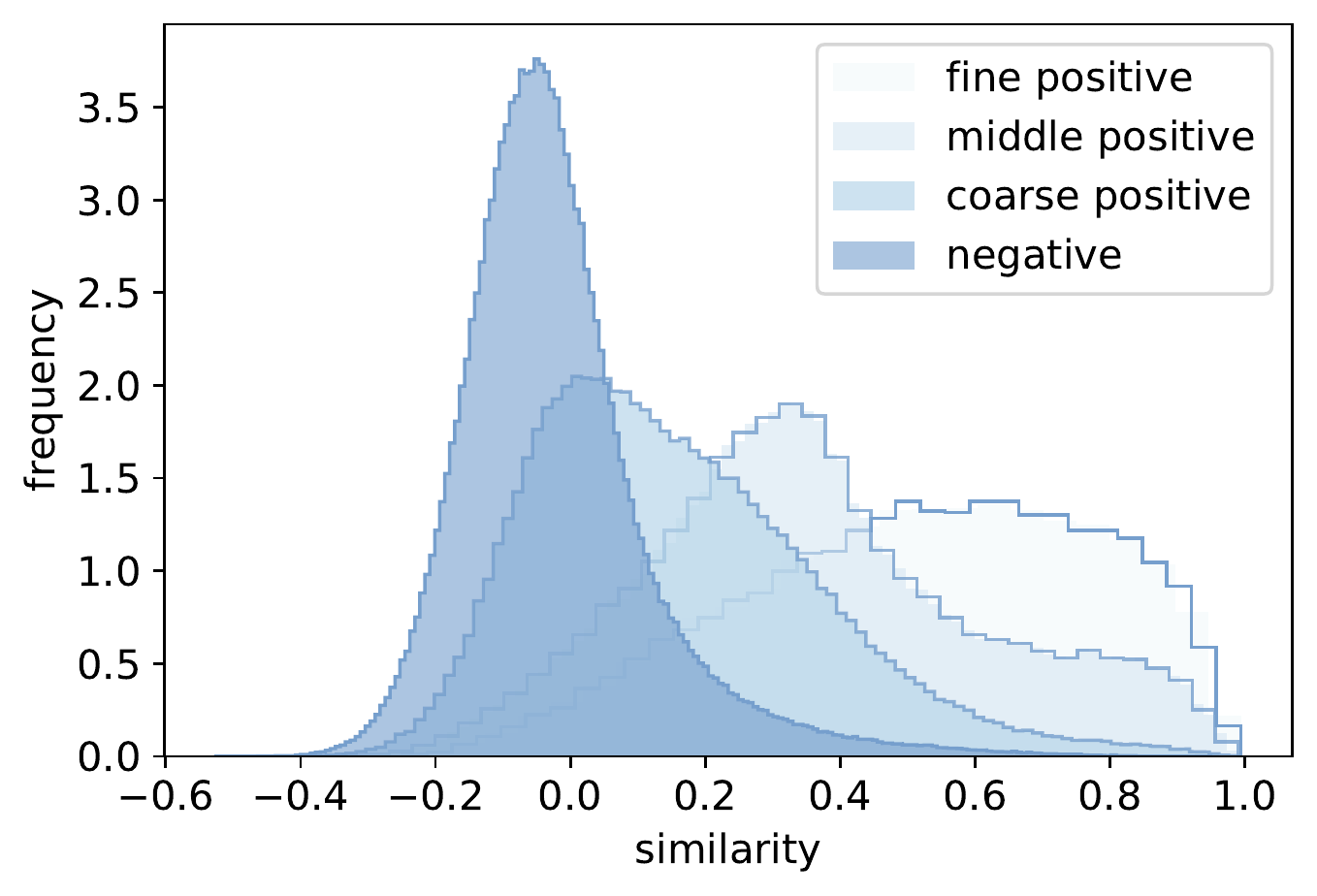}
	\caption{The distribution of similarity scores indicates conflicts between different scales.}
	\vspace{-4mm}
	\label{fig: conflict}
\end{figure}

We investigate the mutual conflict in multi-scale learning baseline, as well as the reason for the superiority of CSL.


\textbf{Mutual conflict between different scales.} During the multi-scale training, we record both the positive similarity scores and the negative similarity scores under three scales in Fig. \ref{fig: conflict}. Overall, the similarity scores of ``fine positive'', ``middle postive'', ``coarse positive'' and ``negative'' pairs are naturally sorted in a descending order. It is because, some negative pairs under the fine scale are actually positive pairs under the middle / coarse scale. When the baseline enforces between-class discrepancy under the fine scale, it tries to decreases their similarity scores, which consequentially decreases the positive similarity scores under the middle / coarse scale. In a word, in the multi-scale learning baseline, the between-class discrepancy in the fine scale compromises the within-class compactness in the middle / coarse scales, and vice versa. We thus conclude that the mutual suppression between different scales hinders the multi-task learning baselines. 

\begin{figure}[t!]
	\centering
	\includegraphics[width=1\linewidth]{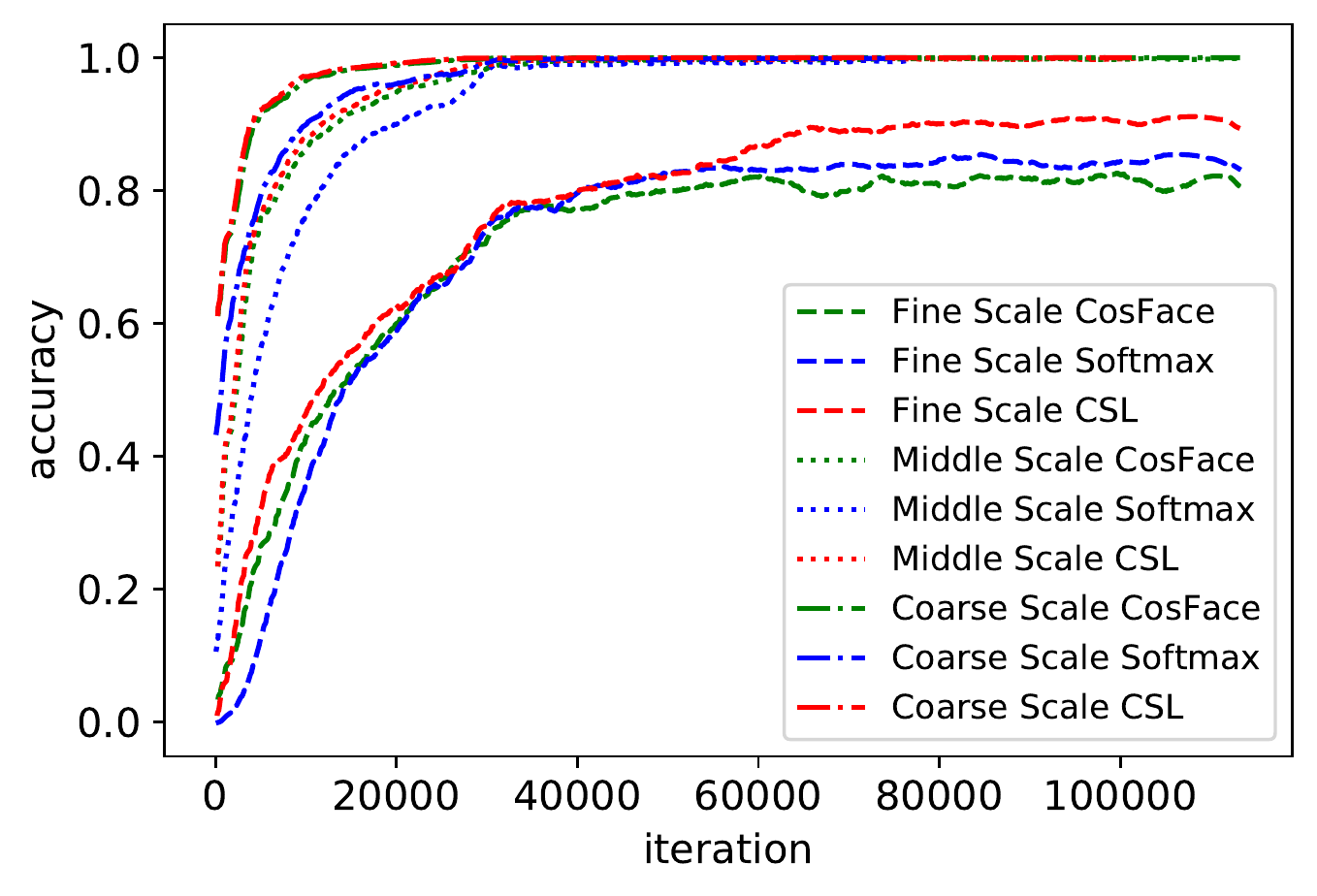}
	\caption{Comparison between the training process of softmax, cosface and CSL on DyML-Animal. Compared with the multi-scale learning baselines, CSL obtains faster convergence and higher classification accuracy on the training set, because it alleviates the mutual conflict between different semantic scales.}
	\label{fig: train_acc}
\end{figure}

\textbf{CSL alleviates the mutual conflict.}
We compare the training process of softmax baseline, Cosface baseline and CSL in Fig. \ref{fig: train_acc}. Specifically, we record the classification accuracy under all the three semantic scales and make the following three observations:

First, both the softmax and the cosface achieves relatively low classification accuracy under the fine scale. After convergence, they achieve 80.3\% and 78.4\% classification accuracy under the fine scale, respectively. It further evidences the mutual conflict phenomenon. Second, comparing CSL against softmax and cosface, we find that CSL facilitates faster and better convergence, especially under the fine scale. After convergence, CSL achieves 84.1\% classification accuracy under the fine scale. It validates that CSL alleviates the mutual conflict. Third, comparing the training process of a same method, we find that in DyML, the convergence under the fine scale is hard to achieve, while the convergence under the middle and raw scales are relatively easier. It is because the fine-grained visual concepts are inherently harder to recognize.

\textbf{Compatibility to both classification-based and pair-based training manner.} We note that CSL is compatible to both the classification-based and pair-based training manner. In Table \ref{tab: evaluation}, CSL adopts the classification training manner. We further compare three training manners, \ie, the classification-based training, pair-based training and the joint training for CSL. The results are shown in Table \ref{tab: ablation}. We observe that CSL trained through classification surpasses its pair-based counterpart, and the joint training further brings incremental improvement. Considering that joint training doubles the hyper-parameters and the improvement is slight, we recommend classification training for CSL. 

\begin{table}
\centering
\setlength{\tabcolsep}{13pt}
\begin{tabular}{lccc}
\hline
             & ASI & mAP & R@1  \\ \hline
CSL (Pair)         & 20.2   & 10.9   & 18.2     \\ 
CSL (Cls)          & 23.0  & 12.1   & 25.2    \\ 
CSL (Cls+Pair)     & 23.7   & 12.6   & 26.1     \\ \hline
\end{tabular}
\caption{Comparison between different training manners for CSL. We present the overall accuracy on DyML-Vehicle.}
\label{tab: ablation} 
\end{table}


\section{Conclusion}

In this paper, we introduce the concept of ``dynamic range'' from real-world metric tools to deep metric for visual recognition. It endows a single metric with scalability to accommodate multiple semantic scales. Based on dynamic range, we propose a new task named Dynamic Metric Learning, construct three datasets (DyML-Vehicle, DyML-Animal and DyML-Product), benchmark these datasets with popular metric learning methods, and design a novel method. 


\section*{Acknowledgement}
This research was supported by China's ``scientific and technological innovation 2030 - major projects'' (No. 2020AAA0104400).

{\small
\bibliographystyle{ieee_fullname}
\bibliography{egbib}
}

\end{document}